\documentclass[conference]{IEEEtran}
\usepackage[hidelinks]{hyperref} 
\usepackage{orcidlink}
\IEEEoverridecommandlockouts
\usepackage{cite}
\usepackage{amsmath,amssymb,amsfonts}
\usepackage{algorithmic}
\usepackage{graphicx}
\usepackage{textcomp}
\usepackage{xcolor}
\usepackage{subcaption}
\usepackage{placeins}
\usepackage{float}
\usepackage{tabularx} 
\usepackage{booktabs}
\usepackage[autostyle=false, style=english]{csquotes}

\usepackage{blindtext}

\usepackage{hyperref}

\usepackage{tikz}
\usetikzlibrary{positioning, decorations.pathreplacing, calligraphy, shapes.geometric, arrows.meta, shapes, bending, calc}

\MakeOuterQuote{"}
\def\BibTeX{{\rm B\kern-.05em{\sc i\kern-.025em b}\kern-.08em
    T\kern-.1667em\lower.7ex\hbox{E}\kern-.125emX}}

\begin{document}
\title{Deep Learning-Based Anomaly Detection\\in Spacecraft Telemetry on Edge Devices
}

\author{\IEEEauthorblockN{Christopher Goetze~\orcidlink{0009-0001-2465-5701}}
\IEEEauthorblockA{
\textit{IU International University}\\
\textit{of Applied Sciences}\\
Germany \\
christopher.goetze@iu-study.org \\
\href{https://orcid.org/0009-0001-2465-5701}{ORCID:0009-0001-2465-5701}}
\and
\IEEEauthorblockN{Tim Schlippe~\orcidlink{0000-0002-9462-8610}}
\IEEEauthorblockA{
\textit{IU International University}\\
\textit{of Applied Sciences}\\
Germany \\
tim.schlippe@iu.org \\
\href{https://orcid.org/0000-0002-9462-8610}{ORCID:0000-0002-9462-8610}}
\and
\IEEEauthorblockN{Daniel Lakey~\orcidlink{0000-0002-8198-7892}}
\IEEEauthorblockA{
\textit{IU International University}\\
\textit{of Applied Sciences}\\
Germany \\
daniel.lakey@iu-study.org \\
\href{https://orcid.org/0000-0002-8198-7892}{ORCID:0000-0002-8198-7892}}
}

\maketitle

\begin{abstract}
Spacecraft anomaly detection is critical for mission safety, yet deploying sophisticated models on-board presents significant challenges due to hardware constraints. This paper investigates three approaches for spacecraft telemetry anomaly detection---\textit{forecasting \& threshold}, \textit{direct classification}, and \textit{image classification}---and optimizes them for edge deployment using multi-objective neural architecture optimization on the European Space Agency Anomaly Dataset. Our baseline experiments demonstrate that \textit{forecasting \& threshold} achieves superior detection performance (92.7\% Corrected Event-wise F$_{0.5}$-score (CEF$_{0.5}$))~\cite{kotowski2024europeanspaceagencybenchmark} compared to alternatives. Through \textit{Pareto-optimal} architecture optimization, we dramatically reduced computational requirements while maintaining capabilities---the optimized \textit{forecasting \& threshold} model preserved 88.8\% CEF$_{0.5}$ while reducing RAM usage by 97.1\% to just 59 KB and operations by 99.4\%. Analysis of deployment viability shows our \textit{optimized} models require just 0.36-6.25\% of CubeSat RAM, making on-board anomaly detection practical even on highly constrained hardware. This research demonstrates that sophisticated anomaly detection capabilities can be successfully deployed within spacecraft edge computing constraints, providing near-instantaneous detection without exceeding hardware limitations or compromising mission safety.
\end{abstract}


\section{Introduction}

Space missions rely critically on the health and operational status of spacecraft systems, making anomaly detection in telemetry data essential for mission success and longevity~\cite{Hundman.2018, lentaris2018high}. Modern spacecraft generate vast amounts of telemetry data---thousands of parameters sampled at frequencies ranging from seconds to minutes---creating an overwhelming volume of information that cannot be manually monitored \cite{Baireddy.2021}. This challenge is compounded by the extreme rarity of actual anomalies, with a comprehensive study of seven spacecraft over a decade identifying fewer than 200 critical anomalies~\cite{lutz2004empirical}.

Traditional ground-based anomaly detection approaches process telemetry data after transmission to Earth, introducing potentially critical delays ranging from seconds for low-Earth orbit missions to hours for deep space missions~\cite{lentaris2018high}. These delays can be substantially longer if there are significant periods when the spacecraft is outside of ground-station visibility \cite{Dyer2018,Woods2006}, potentially compromising mission safety when immediate intervention is required \cite{furano2020towards}. Moreover, bandwidth limitations often necessitate downsampling or selective transmission of telemetry data, potentially obscuring subtle anomalies \cite{furano2020towards}.

On-board anomaly detection offers a compelling alternative by analyzing telemetry data directly on the spacecraft, enabling near-instantaneous detection and response to critical events. This approach minimizes communication delays, conserves limited downlink bandwidth, and allows spacecraft to implement autonomous recovery actions when Earth communication is unavailable \cite{Hundman.2018,Woods2006}. However, deploying sophisticated anomaly detection systems on spacecraft presents significant challenges due to the severe computational constraints of space-grade hardware \cite{lentaris2018high}.

Space-qualified computing systems typically offer just a fraction of the processing power, memory, and storage available in terrestrial environments \cite{furano2020towards}. Additionally, these systems must operate reliably within tight power budgets and withstand harsh radiation environments \cite{lentaris2018high}. These constraints preclude the direct deployment of state-of-the-art anomaly detection models developed for ground-based applications, necessitating specialized approaches for edge implementation.

In this paper, we investigate three distinct deep learning-based approaches for spacecraft telemetry anomaly detection: 

\begin{itemize}
    \item \textit{forecasting \& threshold}, which predicts normal behavior and identifies deviations;
    \item \textit{direct classification}, which directly labels telemetry segments as normal or anomalous; and 
    \item \textit{image classification}, which transforms time series data into images for visual pattern recognition.
\end{itemize}

We conduct comprehensive evaluations using the European Space Agency Benchmark for Anomaly Detection in Satellite Telemetry (ESA-ADB) \cite{kotowski2024europeanspaceagencybenchmark}, containing real operational data from three distinct missions with expert-annotated anomalies.

Beyond comparing model performance, we specifically address the critical challenge of deploying these solutions on space-grade edge devices. We employ neural architecture optimization techniques to systematically reduce model size and computational requirements while preserving detection accuracy. This optimization process leverages multi-objective frameworks that balance performance against hardware constraints, enabling deployment on severely resource-constrained spacecraft computing systems.

Our research makes three primary contributions: 
\begin{itemize}
    \item a comprehensive comparison of deep learning approaches for spacecraft telemetry anomaly detection using real operational data; 
    \item a systematic methodology for optimizing these models for deployment on space-grade edge devices; and 
    \item empirical evidence demonstrating that sophisticated anomaly detection can be achieved within the stringent computational constraints of spacecraft hardware.
\end{itemize}

The remainder of this paper is organized as follows: Section~\ref{Related work} reviews related work in anomaly detection, spacecraft telemetry datasets, and neural architecture optimization. Section~\ref{Experimental Setup} details our experimental setup, including the models, dataset, and evaluation metrics. Section~\ref{Experiments and Results} presents our experimental results and discusses the implications for edge deployment. Finally, Section~\ref{Conclusion and Future Work} concludes the paper and outlines future research directions.

\section{Related work}
\label{Related work}

This section examines current research in spacecraft telemetry anomaly detection, neural architecture optimization for resource-constrained environments, and available spacecraft telemetry datasets. 

\subsection{Anomaly Detection in Spacecraft Telemetry}

Anomaly detection in spacecraft telemetry is critical for mission safety and success \cite{Heras.2014,Hundman.2018,Pilastre.2020,Baireddy.2021,Baireddy.2023}. Three primary approaches have emerged in this domain: \textit{forecasting \& threshold}, \textit{direct classification}, and \textit{image classification}. Recent surveys like~\cite{zamanzadeh2023} provide comprehensive overviews of deep learning techniques for time series anomaly detection across multiple domains, highlighting the increasing importance of these methods for critical systems monitoring.

\subsubsection{Forecasting \& Threshold}\label{lab:forecasting_thresh}

The \textit{forecasting \& threshold} approach predicts future values based on historical data and identifies anomalies when actual values deviate beyond a specified threshold. The model forecasts values based on learned patterns from previous timesteps, then compares these predictions with observations to quantify anomalousness. This method requires training on normal data to establish baseline behavior patterns. Various models have been adapted for this purpose, including statistical methods like AutoRegressive Integrated Moving Average (ARIMA)~\cite{Ihler2006} and deep learning architectures such as Telemanom's Long Short-Term Memories (LSTMs)~\cite{Hundman.2018} and DeepAnT's Convolutional Neural Networks (CNNs)~\cite{Munir2019}. \cite{Yang2021}~demonstrated significant improvements in spacecraft anomaly detection by combining LSTM networks with multi-scale detection strategies, validating their approach on NASA benchmark data and Beidou Navigation Satellite hydrogen clock telemetry. Their work highlights how enhanced deep learning architectures can significantly improve operational reliability through early anomaly detection.

The effectiveness of this approach depends heavily on threshold selection. \cite{Hundman.2018}~proposed an "unsupervised and nonparametric anomaly thresholding approach" that dynamically learns appropriate error thresholds for each telemetry channel. While over 158 time series anomaly detection models existed as of 2022~\cite{WenigEtAl2022TimeEval}, many are designed for univariate data~\cite{Schmidl.2022}, limiting their applicability to the multivariate context of spacecraft telemetry.

\subsubsection{Direct Classification}\label{lab:time_series_classif}

\textit{Direct classification} approaches label time points or windows directly as normal or anomalous without intermediate forecasting steps. This method circumvents the challenges of complex threshold tuning and forecasting inaccuracies~\cite{Harrell_2017}. Deep learning architectures like CNNs~\cite{wang2016time,IsmailFawaz2019,Koh2021} and RNNs~\cite{Karim2018,Altaf2018} have demonstrated promising results in capturing temporal dynamics for classification tasks. As detailed in the comprehensive survey by~\cite{zamanzadeh2023}, these \textit{direct classification} approaches have gained significant traction across multiple domains due to their ability to learn complex patterns directly from data without manual feature engineering. Recent research by~\cite{lakey2024anomaly} indicates that \textit{direct classification} provides statistically significant improvements over traditional forecasting methods for spacecraft telemetry.

\subsubsection{Image Classification}\label{lab:lr_image_encoding}

The \textit{image classification} approach transforms time series data into images before applying classification algorithms. This leverages the exceptional performance of CNNs in image recognition~\cite{Sharma2018,Staar2019}. A prominent transformation technique is the Gramian Angular Field (GAF)~\cite{Sreeram1994,Wang2014EncodingTS,Wang.2015}, which encodes time series into polar coordinates and then into a two-dimensional representation using Gram matrices~\cite{Vitry_2018}.

GAF transformations preserve temporal dependencies by encoding time into the geometry of the matrix \cite{Vitry_2018}. This approach has been successfully applied in financial forecasting~\cite{Barra.2020} and general anomaly detection~\cite{Sastry.2019}. While \cite{lakey2024anomaly} found \textit{image classification} less effective than \textit{direct classification} for spacecraft telemetry, they suggest it offers promising directions for future research in this domain.

\subsection{Neural Architecture Search and Optimization}
\label{Neural Architecture Search and Optimization}

Deploying deep learning models for spacecraft anomaly detection on edge devices requires addressing severe hardware constraints \cite{lentaris2018high, furano2020towards}. Unlike terrestrial systems that can be upgraded, spacecraft computing resources remain fixed throughout the mission lifetime, making efficient resource utilization critical \cite{lentaris2018high}.

CubeSat~\cite{SatCatalog.2021} and OPS-SAT~\cite{Evans.2016,Evans.2017} are two prevalent on-board computing platforms utilized in contemporary spacecraft, each catering to different operational scopes and resource capabilities. CubeSats are compact, standardized satellites designed for cost-effective, modular deployment in space research missions. A typical CubeSat features modest computational resources, including a PIC32MZ M14K processor operating at speeds of up to 200 MHz, along with 16 MB of Random Access Memory (RAM) and 64 MB of flash memory~\cite{SatCatalog.2021}. In contrast, ESA’s OPS-SAT serves as an advanced, flexible experimentation platform equipped with more robust hardware, making it suitable for demanding computational tasks. Specifically, OPS-SAT incorporates an Altera Cyclone V System-on-Chip~(SoC) featuring an ARM dual-core 800 MHz Cortex-A9 processor, complemented by 1 GB DDR3 RAM and an external 8 GB industrial-grade SD card storage~\cite{Evans.2017}. Given these limited computational resources, particularly in the CubeSat environment, deploying an anomaly detection model necessitates careful optimization. Minimizing the model's resource footprint ensures that sufficient processing power is available for critical spacecraft subsystems, thereby maintaining operational reliability.

A fundamental concept in balancing competing objectives like model accuracy and computational efficiency is the \textit{Pareto front} approach to multi-objective optimization \cite{Jin.2008}. \textit{Pareto-optimal} represents a set of non-dominated solutions where a solution $A$ dominates solution $B$ if $A$ is at least as good as $B$ in all objectives and strictly better in at least one. Solutions on this front cannot be improved in any objective without degrading performance in another, representing optimal trade-offs between conflicting goals \cite{Jin.2008}. This approach is particularly valuable for spacecraft applications where designers must balance detection accuracy against strict computational limitations, though it does not prescribe a single optimal solution but rather presents a set of efficient alternatives from which mission designers can select based on specific deployment constraints.

\textit{Neural architecture search} and \textit{neural architecture optimization} represent two approaches to this challenge. \textit{Neural architecture search} focuses on discovering entirely new architectures, exemplified by MicroNAS \cite{liu2024micronas} and $\mu$NAS \cite{banbury2021micronets}. While $\mu$NAS has shown impressive results---improving accuracy by up to 4.8\% while reducing memory usage by 4--13×---it requires substantial computational resources (23 GPU days for CIFAR10) and generates entirely new, unproven architectures.

For spacecraft applications, \textit{neural architecture optimization} often provides a more practical approach by optimizing proven architectures rather than creating new ones, which is particularly important where reliability is critical \cite{george2020spacecraft}. \textit{Multi-objective optimization} frameworks like Optuna \cite{akiba2019optuna} are especially valuable for this task. Optuna offers intelligent sampling algorithms that simultaneously consider both model accuracy and computational constraints, making it ideal for spacecraft applications with strict resource limitations.

Effective visualization and profiling tools are essential components of the architecture optimization process. Two notable examples are the CNN Analyzer and the MLTK Model Profiler. \cite{brockmann2024optimizing}~utilized their CNN Analyzer to provide layer-wise insights into how different scaling factors affect model size, peak memory usage, and inference time. This dashboard-based tool enabled them to systematically optimize model architecture scaling factors, resulting in a 20.5\% reduction in model size while simultaneously increasing accuracy by 4.0\%. 
Similarly, the MLTK Model Profiler \cite{siliconlabs2023mltk} serves the same fundamental purpose of visualizing and analyzing model efficiency, but with features specifically designed for space-grade embedded systems. Both tools enable developers to identify optimization opportunities by providing detailed breakdowns of resource consumption, but the MLTK Model Profiler offers additional integration with spacecraft hardware development workflows, making it particularly suitable for our application in spacecraft telemetry anomaly detection.

In our work, we combined Optuna's optimization capabilities with the detailed insights provided by the MLTK Model Profiler to optimize our anomaly detection models for spacecraft edge devices. Utilizing the \textit{Pareto-optimal} approach for multi-objective optimization \cite{Jin.2008}, we systematically balanced model performance against computational constraints. This \textit{neural architecture optimization} approach enabled us to develop sophisticated anomaly detection capabilities that operate within the strict limitations of on-board hardware, making it possible to deploy effective deep learning models for on-board analysis without exceeding the stringent memory and computational constraints of space-grade computing systems.

\subsection{Spacecraft Telemetry Data}
\label{Spacecraft Telemetry Data}

Spacecraft telemetry data presents unique challenges for anomaly detection due to its high dimensionality and the rarity of actual anomalies. Modern spacecraft generate data from tens of thousands of telemetry channels~\cite{lakey2024presentation}, producing volumes that exceed human monitoring capabilities \cite{Baireddy.2021, yairi2014evaluation}. Spacecraft are designed to be robust and fault-tolerant, with extensive testing to minimize anomalies \cite{Fortescue_Swinerd_Stark_2011}, resulting in extremely imbalanced datasets where anomalous events are scarce---a study across seven spacecraft over more than a decade identified fewer than 200 critical anomalies \cite{lutz2004empirical}. 

A significant obstacle in spacecraft anomaly detection research has been the scarcity of publicly available datasets. The SMAP/MSL dataset from \cite{Hundman.2018} has become a common benchmark, containing 82 multivariate telemetry channels from the Soil Moisture Active Passive spacecraft and Mars Science Laboratory with approximately 100 labeled anomalies. However, this dataset has recognized limitations, including anonymized channel identifiers and normalized values \cite{Wu.2021,lakey2024anomaly}. 

More recently, the European Space Agency Benchmark for Anomaly Detection in Satellite Telemetry (ESA-ADB) has emerged as a valuable resource for researchers \cite{kotowski2024europeanspaceagencybenchmark}. It represents the first large-scale, real-life satellite telemetry dataset with curated anomaly annotations from three European Space Agency missions. Developed through an 18-month collaboration between Airbus Defence and Space, KP Labs, and ESA's European Space Operations Centre, this dataset aims to provide a common benchmark for researchers across academia, research institutes, space agencies, and industry. The ESA-ADB is part of the broader Artificial Intelligence for Automation (A²I) Roadmap initiative launched in 2021 to leverage AI for automating space operations. In this work, we utilize the ESA-ADB for our experiments, leveraging its comprehensive real-world telemetry data to develop and validate our edge-optimized anomaly detection models.

\section{Experimental Setup}
\label{Experimental Setup}

\subsection{Forecasting \& Threshold Models}\label{lab:forecasting_thresh_setup}

For our forecasting-based anomaly detection approach, we built upon our previous research \cite{lakey.2024} where we extended the Telemanom framework originally developed by~\cite{Hundman.2018}. While the original implementation relied exclusively on LSTM-based forecasting with a non-parametric thresholding mechanism, our prior work comprehensively evaluated 13~different deep learning architectures as potential forecasting components within this framework.

Our experiments identified XceptionTimePlus as the optimal architecture for spacecraft telemetry forecasting~\cite{lakey2024anomaly}. This model, based on the XceptionTime architecture \cite{rahimian2019xceptiontime} from the \verb|tsai| time series analysis framework \cite{tsai}, leverages convolutional neural networks with depthwise separable convolutions to capture temporal patterns efficiently. This CNN-based approach demonstrated superior performance compared to the original LSTM implementation across multiple telemetry channels.

In the current study, we evaluate this optimized \textit{forecasting \& threshold} system (XceptionTimePlus with non-parametric thresholding) on the ESA-ADB. This approach serves as our \textit{baseline} for comparison against the \textit{direct classification} and \textit{image classification} techniques described in subsequent sections.

\subsection{Direct Classification}

For our \textit{direct classification} approach, we implemented the XceptionTimePlus architecture, a variant of the XceptionTime model \cite{rahimian2019xceptiontime} available in the \verb|tsai| framework \cite{tsai}. We configured the model using the optimal hyperparameters identified in our previous study on the SMAP/MSL dataset~\cite{lakey.2024}. This choice is well-supported by existing literature, as XceptionTime has demonstrated superior performance in multiple non-spacecraft anomaly detection studies \cite{bickmann2023post,Nazir2023} and shows excellent capability in handling multivariate time series data, making it particularly suitable for spacecraft telemetry analysis.

Our implementation leveraged the Python time series classification framework \verb|tsai| \cite{tsai}, executed within Kaggle Notebooks\footnote{https://www.kaggle.com/docs/notebooks} utilizing an Nvidia P100 GPU\footnote{https://www.kaggle.com/docs/efficient-gpu-usage}. We developed a comprehensive classification pipeline that:

\begin{itemize}
    \item Configures the appropriate loss function, optimizer, and class weights
    \item Instantiates the XceptionTimePlus model with optimal parameters
    \item Performs direct binary classification (normal/anomalous) on time windows (224 time points as in~\cite{lakey2024anomaly})
    \item Evaluates performance metrics including true positives, false positives, false negatives, and Corrected Event-wise F$_{0.5}$-score~\cite{kotowski2024europeanspaceagencybenchmark}
\end{itemize}

The same pipeline architecture, with an additional transformation stage, was later adapted for our \textit{image classification} experiments, providing a consistent experimental framework across all our anomaly detection approaches.

\subsection{Image Classification}

Image encoding of time series data represents an innovative approach to anomaly detection that leverages the power of convolutional neural networks by transforming temporal data into spatial representations \cite{Jiang2022, wang2016time}. For our implementation, we utilized the Gramian Angular Field (GAF) transformation from the \verb|pyts| framework \cite{JMLR:v21:19-763} to convert telemetry time series into image-based representations suitable for classification.

We developed a custom PyTorch-accelerated transformation module implementation\footnote{https://github.com/chris-official/PyTorchGAF} that significantly improves upon the original pyts. Our implementation eliminates unnecessary data reshaping and CPU-GPU transfers, performs calculations directly in batched operations on the target device, and uses vectorized operations with Einstein Summation Notation for matrix calculations. This optimization achieved an 80× speed-up with GPU acceleration compared to the original implementation, critically enabling efficient image transformation for our edge deployment targets.

The standard GAF transformation is designed for univariate (single parameter) time series data, presenting a challenge for spacecraft telemetry which is inherently multivariate with multiple interconnected parameters. To address this limitation while preserving the critical contextual relationships between parameters, we implemented a multi-channel approach. This method generates separate GAF images for each parameter within a telemetry channel, creating a ``stack'' of images that collectively preserves the multivariate nature of the data. 

We explored alternative multivariate image transformations such as the recurrence plot method \cite{Eckmann1987} available in \verb|pyts|, but our preliminary experiments yielded poor results with no anomalies detected, leading us to focus exclusively on the stacked GAF approach.

For the classification of these transformed images, we implemented two variants of xResNet34, a \verb|tsai| adaptation of the ResNet34 architecture \cite{he2015deep, He.2018} known for its exceptional performance in \textit{image classification} tasks \cite{Gao2021, Zhuang2022, Gao2023}. Unlike XceptionTimePlus used in our \textit{direct classification} approach, ResNet34 is specifically designed to handle multi-dimensional image inputs, making it ideal for our transformed telemetry data.

We evaluated both an untrained xResNet34 model trained from scratch on our telemetry images and a pre-trained version (referred to as xResNet34$_{\text{(fine-tuned)}}$) from the \verb|PyTorch| collection \cite{pytorchModelsPretrained} that we fine-tuned on our dataset. The fine-tuned approach was expected to provide better performance given the relatively limited number of training samples available for this task \cite{Shahinfar2020}.

Our implementation standardized on $224\times224$ pixel images for the GAF transformations, a resolution widely recognized in the literature as providing an optimal balance of detail and computational efficiency \cite{Simonyan.2014, Howard.2017, Zhang.2022, Richter2021}. This resolution also aligned with the pre-training dimensions of the xResNet34$_{\text{(fine-tuned)}}$ model \cite{He.2015}, potentially enhancing transfer learning effectiveness.

\subsection{ESA-ADB}

For our experiments, we utilized the European Space Agency Benchmark for Anomaly Detection in Satellite Telemetry (ESA-ADB)~\cite{kotowski2024europeanspaceagencybenchmark}, the first large-scale, real-life satellite telemetry dataset with curated anomaly annotations. This dataset represents a significant advancement in spacecraft telemetry analysis, providing researchers with authentic operational data to develop and benchmark anomaly detection models.

The ESA Anomalies Dataset (ESA-AD) comprises telemetry data from three distinct ESA missions, with two of them included in the ESA-ADB. These missions represent diverse spacecraft types, orbits, and operational profiles. The telemetry spans a significant period, with an anonymized duration of approximately 17.5~years in total.

A key strength of this dataset is its comprehensive annotation scheme. The dataset includes 844 annotated events across two missions used in the benchmark, with 148 classified as anomalies.
These events were meticulously labeled by spacecraft operations engineers and machine learning experts, and cross-verified using state-of-the-art algorithms.

The dataset is structured as a multivariate time series with the following characteristics:
\begin{itemize}
    \item Complex characteristics including varying sampling frequencies across time and channels
    \item Data gaps caused by idle states and communication problems
    \item Trends connected with degradation of spacecraft components
    \item Concept drifts related to different operational modes and mission phases
    \item Diverse channel types including physical 
    measurements, categorical status flags, counters, and binary telecommands
\end{itemize}

The dataset contains 176 channels in total (76 for Mission~1 and 100 for Mission~2), with 105 designated as target channels for anomaly detection and 71 as non-target channels to support the detection process. The total volume exceeds 1.55~billion data points, comprising over 7~gigabytes of compressed data.

Unlike previous spacecraft telemetry datasets, the ESA-AD maintains real operational characteristics while addressing privacy concerns through anonymization processes that preserve data integrity. This preservation of authentic data patterns is particularly valuable for anomaly detection research, as it allows evaluation of model performance against the actual complexities and challenges of space operations.

To support reproducible research, the dataset includes standardized train/validation/test splits and a comprehensive evaluation pipeline with new metrics designed specifically for satellite telemetry according to the latest advancements in time series anomaly detection. 

For our experiments, we focused on the lightweight subset of Mission~1 since according to~\cite{kotowski2024europeanspaceagencybenchmark} it is 
\begin{itemize}
    \item suitable for potential on-board applications, challenging for algorithms,
    \item interesting for spacecraft operations engineers, 
    \item relatively easy to visualize and analyze manually and
    \item not strongly dependent on other channels or subsystems.
\end{itemize}

\subsection{Evaluation Metrics}

To evaluate the performance of our anomaly detection models, we adopted the Corrected Event-wise F$_{0.5}$-score (CEF$_{0.5}$)~\cite{kotowski2024europeanspaceagencybenchmark} as recommended in the ESA-ADB~\cite{kotowski2024europeanspaceagencybenchmark}. This specialized metric addresses several critical challenges inherent to spacecraft telemetry anomaly detection that are not adequately captured by traditional point-wise metrics.

Unlike conventional F1 scores that treat each data point independently, the CEF$_{0.5}$ operates at the event level, addressing the temporal structure of spacecraft anomalies. This approach recognizes that anomalies in spacecraft telemetry typically manifest as continuous segments rather than isolated points, and a model's practical utility depends on detecting these anomalous events rather than maximizing the number of correctly classified individual data points. This builds on the anomaly-based scoring approach followed by~\cite{Hundman.2018,lakey.2024}.

The metric incorporates three key innovations tailored to spacecraft telemetry evaluation:

\begin{enumerate}
    \item \textbf{Event-based evaluation:} Rather than treating each time point independently, the metric first converts both predicted and ground truth anomalies into discrete events. An event is defined as a continuous sequence of anomalous points, where events separated by less than a predefined tolerance window (60 seconds in our implementation) are merged into a single event.
    
    \item \textbf{Directional tolerance:} The metric implements an adjustable timing tolerance that accounts for early detection, recognizing that in operational scenarios, detecting an anomaly slightly before its actual occurrence can be as valuable as detecting it precisely when it occurs.
    
    \item \textbf{Precision emphasis:} The CEF$_{0.5}$ variant weights precision higher than recall (by a factor of 2), reflecting the operational reality that false alarms are particularly costly in spacecraft operations, where they may trigger unnecessary interventions that consume limited resources.
\end{enumerate}

CEF$_{0.5}$ is based on precision and recall which are determined at the event level. An event is considered correctly detected if there is at least one predicted anomalous point that temporally overlaps with the ground truth event after applying the tolerance window.

This metric provides a more operationally relevant assessment of anomaly detection performance in spacecraft telemetry than traditional point-wise metrics, aligning our evaluation with the practical needs of mission operations where timely event detection with minimal false alarms is paramount.

\subsection{Optimization for Edge Devices}
We employed multi-objective optimization to balance the critical tradeoff between model accuracy and computational efficiency for spacecraft edge deployment. Rather than using CEF$_{0.5}$ as our accuracy metric, we selected validation loss as our primary performance indicator since it provides a more stable optimization target that does not depend on threshold selection, while also being computationally efficient to calculate during training. 

For model complexity assessment, we utilized Multiply-Accumulate operations (MACs) as a proxy metric, which has become a standard practice in hardware-aware neural architecture search \cite{banbury2021micronets, benmeziane2021comprehensive, garavagno2024affordable, hsu2018monas, yang2023neural}. The $\mu$NAS framework authors observed a strong linear correlation between MACs and actual microcontroller unit inference latency, validating its use as an accurate approximation for deployment performance \cite{liberis2020unas}. While some researchers have criticized MACs as potentially inaccurate \cite{benmeziane2021comprehensive, king2025micronas, tan2019mnasnet, wu2018fbnet}, more precise alternatives typically require physical access to target hardware and introduce substantial evaluation overhead, making them impractical for our optimization process. Direct measurements on target devices would significantly slow the optimization cycle \cite{benmeziane2021comprehensive, king2025micronas}. 

Our optimization specifically targeted space-qualified computing systems which typically offer just 10-100x less processing power, memory, and storage compared to their terrestrial counterparts \cite{lentaris2018high}. These substantial constraints are further compounded by radiation hardening requirements that often necessitate the use of older, more robust but less powerful semiconductor technologies \cite{furano2020towards}. Unlike terrestrial edge devices, spacecraft computing hardware cannot be upgraded once deployed, making efficient resource utilization particularly critical throughout the mission lifetime. 

To accelerate our optimization process while preserving data diversity, we trained on 2~million samples from the 7~million available training samples, using sample weights to ensure exposure to all anomalous sequences. We standardized training at 2,048 steps per model configuration to ensure fair comparisons, with evaluations performed every 256 steps on the validation dataset (approximately 265,000 samples)---substantially more efficient than using the 7 million sample test dataset while avoiding potential test set overfitting. 

Our optimization framework utilized Optuna's Tree-structured Parzen Estimator (TPE) sampler to efficiently balance exploration and exploitation of the search space \cite{akiba2019optuna}. Each optimization run adjusted key architectural parameters including layer counts, filter sizes, and activation functions to achieve optimal performance within strict computational constraints. We fixed learning rate and batch size parameters to ensure performance differences resulted from architectural improvements rather than training hyperparameters. After training more than 50 model configurations for each approach, we identified \textit{Pareto-optimal} configurations following the principles established by \cite{Jin.2008}, where a solution is considered \textit{Pareto-optimal} if it cannot be improved in one objective (accuracy) without degrading another (computational efficiency). This approach yielded a frontier of non-dominated solutions that represented the best possible tradeoffs between our competing objectives, allowing mission designers to select implementations according to specific hardware constraints. Importantly, we pruned configurations exceeding our \textit{baseline} model size to focus computational resources on promising lightweight alternatives suitable for spacecraft deployment.

For comprehensive model analysis, we utilized the MLTK Model Profiler to generate detailed insights into layer-wise computational demands and memory requirements \cite{siliconlabs2023mltk}. This specialized profiler is particularly valuable for space applications as it provides accurate estimates of peak memory usage, MACs distribution across layers, and overall model efficiency on resource-constrained hardware. The insights gained from this profiling guided our strategic decisions in architecture modification, helping identify specific layers and components that could be optimized without compromising detection performance.

\begin{table*}[t]
    \centering
    \caption{Performances of Baseline and Optimized Anomaly Detection Systems}
    \label{tab:metric_results}
    \begin{tabularx}{\textwidth}{l *{6}{>{\centering\arraybackslash}X}}
        \toprule
         & \multicolumn{3}{c}{Baseline} & \multicolumn{3}{c}{Optimized} \\
        \cmidrule(lr){2-4} \cmidrule(lr){5-7}
        Metric & \textit{Direct classification} & \textit{Image classification} & \textit{Forecasting \& threshold} & \textit{Direct classification} & \textit{Image classification} & \textit{Forecasting \& threshold} \\
        \midrule
        Event-wise Precision (\%) & 95.3 & 50.2 & 99.8 & 91.1 & 99.1 & 99.9 \\
        Event-wise Recall (\%)   & 36.9 & 43.1 & 72.3 & 16.9 & 32.3 & 61.6 \\
        Event-wise F$_{0.5}$ (\%)     & 72.4 & 48.6 & \textbf{92.7} & 48.5 & 70.1 & 88.8 \\
        \midrule
        Parameters    & 310k & 1,254k & 317k & \textbf{19k} & 128k & 29k \\
        MACs         & 56,538k  & 959,367k  & 56,735k  & 907k    & 45,000k  & \textbf{354k}    \\
        RAM (KB)      & 1,606   & 8,193    & 2,043   & 122    & 1,024   & \textbf{59}     \\
        ROM (KB)      & 1,268   & 4,902    & 1,294   & \textbf{149}    & 508     & 166    \\
        \bottomrule
    \end{tabularx}
\end{table*}

\section{Experiments and Results}
\label{Experiments and Results}

\subsection{Results of Forecasting \& Threshold, Direct Classification and Image Classification}

Table \ref{tab:metric_results} presents the comparative performance of our \textit{baseline} and \textit{optimized} anomaly detection approaches. Among the \textit{baseline} models, the \textit{forecasting \& threshold} approach demonstrated superior detection capability with the highest CEF$_{0.5}$ of 92.7\%, substantially outperforming both \textit{direct classification} (72.4\%) and \textit{image classification} (48.6\%). This performance advantage stems primarily from its exceptional precision (99.8\%) combined with reasonable recall (72.3\%), indicating that when this approach identifies an anomaly, it is almost certainly correct---a particularly valuable characteristic for spacecraft operations where false alarms can trigger costly interventions.

\textit{Direct classification} achieved high precision (95.3\%) but struggled with recall (36.9\%), suggesting that while it rarely generates false alarms, it misses a significant portion of actual anomalies. This limitation likely stems from the approach's inability to leverage temporal context beyond its fixed window size, making subtle deviations difficult to detect. Despite this limitation, it is worth noting that our \textit{direct classification} implementation still outperforms most benchmark models~\cite{kotowski2024europeanspaceagencybenchmark} on the ESA-ADB in terms of CEF$_{0.5}$, demonstrating its viability as an anomaly detection approach. This precision-oriented performance aligns with spacecraft operators' preferences, who typically consider false alarms more problematic than missed anomalies, as they can rely on spacecraft autonomy for basic protection while avoiding unnecessary interventions. This operational reality reinforces existing research indicating that minimizing false positives is a critical consideration in spacecraft operations~\cite{Hundman.2018,kotowski2024europeanspaceagencybenchmark}.

The \textit{image classification} approach, despite its theoretical advantages in spatial pattern recognition, performed surprisingly poorly with \textit{baseline} models, showing the lowest CEF$_{0.5}$ (48.6\%) and precision (50.2\%). This underperformance may be attributed to information loss during the GAF transformation process, where temporal relationships are encoded into spatial patterns that may not be effectively interpreted by the model without extensive training data. Interestingly, however, the recall of 43.1\% slightly exceeds that of the \textit{direct classification} approach, suggesting potential complementarity between these methods.

From a computational perspective, both \textit{direct classification} and \textit{forecasting \& threshold} demonstrated comparable resource requirements, with approximately 310k-317k parameters and 56M-57M MACs. Despite their similar computational footprints, the \textit{forecasting \& threshold} approach achieved significantly better detection performance, suggesting superior algorithmic efficiency for spacecraft telemetry analysis. The \textit{image classification} approach, with over 1.25 million parameters and 959M MACs, proved substantially more resource-intensive while delivering the poorest detection results, making it the least efficient option for space-constrained deployment scenarios.

Memory utilization patterns further highlight the implementation challenges of these approaches. \textit{Image classification} required approximately 8,193 KB of RAM---four times more than the other approaches---due to the memory-intensive nature of image transformation and processing. This memory footprint would make deployment impractical on many spacecraft computing platforms that typically offer limited RAM resources. \textit{Direct classification} and \textit{forecasting \& threshold} demonstrated similar memory profiles (1,606 KB and 2,043 KB of RAM, respectively), aligning with their comparable parameter counts.

These \textit{baseline} results indicate that \textit{forecasting \& threshold} offers the best combination of detection capability and computational efficiency for spacecraft telemetry anomaly detection. Its superior CEF$_{0.5}$, coupled with reasonable resource requirements, makes it particularly well-suited for spacecraft deployment scenarios where both accurate anomaly detection and computational efficiency are critical. \textit{Direct classification} presents a viable alternative with good precision but limited recall, while \textit{baseline} \textit{image classification} appears unsuitable for spacecraft deployment without significant optimization, due to its poor detection performance and excessive resource requirements.

\subsection{Optimization for Edge Devices}

\subsubsection{Pareto-Optimal Architecture Selection}

\begin{figure}[h!]
    \centering\includegraphics[width=\columnwidth]{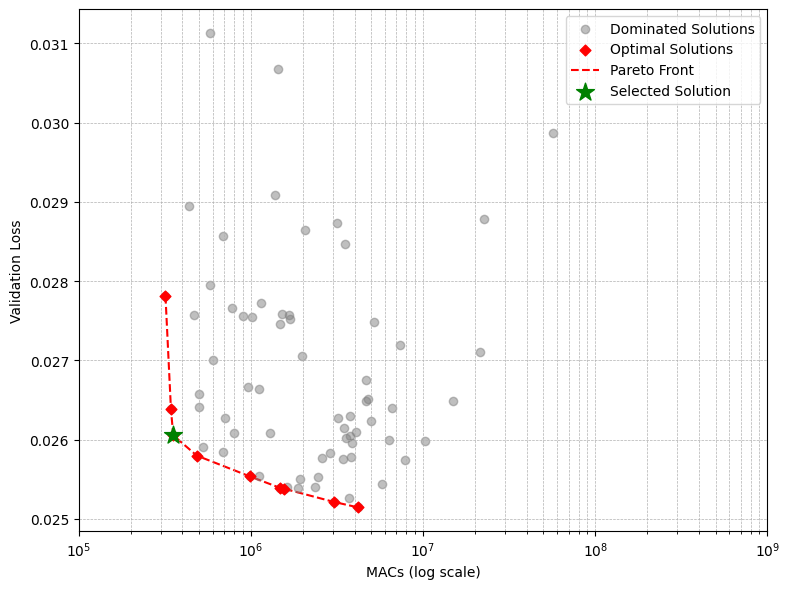}
    \caption{Pareto Front Plot for \textit{Forecasting \& Threshold}.}
    \label{fig:pareto_forecast}
\end{figure}

\begin{figure}[h!]
    \centering\includegraphics[width=\columnwidth]{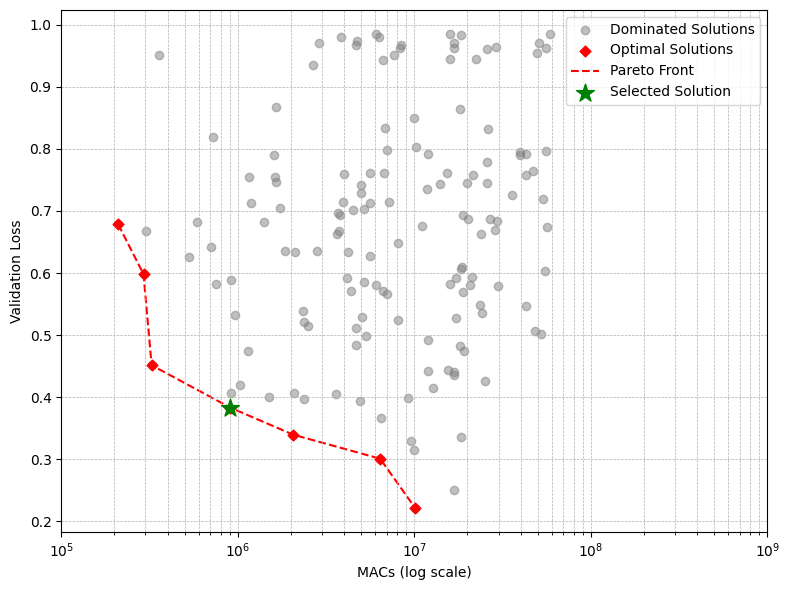}
    \caption{Pareto Front Plot for \textit{Direct Classification}.}
    \label{fig:pareto_direct}
\end{figure}

\begin{figure}[h!]
    \centering\includegraphics[width=\columnwidth]{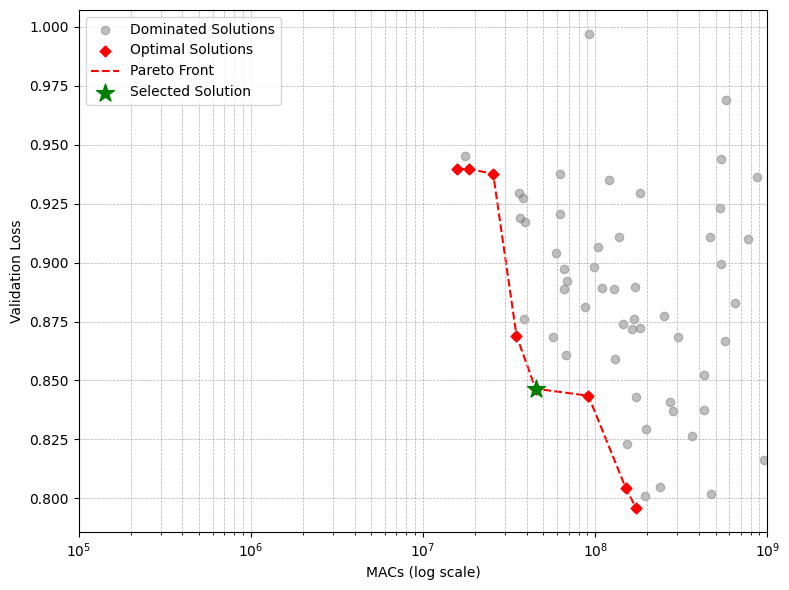}
    \caption{Pareto Front Plot for \textit{Image Classification}.}
    \label{fig:pareto_image}
\end{figure}

The core challenge of deploying deep learning models on spacecraft hardware lies in balancing detection accuracy against severe computational constraints. To systematically explore this tradeoff, we applied multi-objective optimization using the \textit{Pareto-optimal} approach, which identifies solutions where one objective cannot be improved without degrading another. Figures~\ref{fig:pareto_forecast}, \ref{fig:pareto_direct}, and \ref{fig:pareto_image} visualize this optimization process for our three anomaly detection approaches.

In each figure, the horizontal axis represents computational complexity measured in MACs, while the vertical axis shows validation loss, our proxy for model accuracy. The gray dots represent candidate model configurations evaluated during our optimization process. These \textit{dominated} solutions---configurations for which there exists at least one alternative that is superior in at least one objective without being inferior in the other---represent inefficient designs that should be avoided in resource-constrained environments.

The red dots mark the \textit{Pareto-optimal} configurations that form the \textit{Pareto-optimal}---models that achieve the best possible validation loss for a given computational budget or, equivalently, the lowest computational cost for a given level of accuracy. These non-dominated solutions represent the most efficient implementations of each detection approach, providing spacecraft engineers with a range of deployment options calibrated to specific mission constraints.

For our final model selection, we prioritized balanced performance by choosing configurations closest to the origin (marked as a green dot), sometimes referred to as the ``knee point'' of the \textit{Pareto} curve. This selection criterion identifies models that deliver substantial reductions in computational requirements while maintaining acceptable detection performance---a critical consideration for spacecraft deployment where both factors significantly impact mission success.

The \textit{Pareto} plots reveal distinct optimization characteristics across our three detection approaches. The \textit{forecasting \& threshold} approach (Figure~\ref{fig:pareto_forecast}) demonstrates a relatively gradual tradeoff curve, suggesting that detection performance can be maintained even with significant reductions in model complexity. The \textit{direct classification} approach (Figure~\ref{fig:pareto_direct}) exhibits a steeper \textit{Pareto-optimal}, indicating a more sensitive relationship between computational resources and detection capability. The \textit{image classification} optimization (Figure \ref{fig:pareto_image}) shows the most challenging tradeoff landscape, with substantial performance penalties accompanying even moderate reductions in computational resources.

These optimization results informed our final model selection, guiding us toward implementations that maintain critical detection capabilities while operating within the severe computational constraints of spacecraft edge devices. The selected configurations, marked on each \textit{Pareto-optimal}, were subsequently evaluated on the complete test dataset to verify their real-world performance, as detailed in the following sections.

\subsection{Resource Usage in On-Board Computing Environments}

To assess the practical deployability of our \textit{optimized} models, we analyzed their resource requirements in the context of two representative spacecraft computing platforms: the resource-constrained CubeSat with its PIC32MZ M14K processor (200 MHz), 16 MB of RAM and 64 MB of flash memory, and the more capable OPS-SAT experimental platform with an Altera Cyclone V SoC (ARM dual-core 800 MHz Cortex-A9), 1 GB DDR3 RAM and 8 GB industrial-grade SD card storage. Table~\ref{tab:resource_results} quantifies our models' requirements as absolute values and as percentages of available resources on each platform.

The \textit{baseline} implementations demonstrate why direct deployment of sophisticated anomaly detection models on spacecraft hardware has been challenging. While all \textit{baseline} models could technically fit within OPS-SAT's generous memory allocation, they would consume significant resources---particularly the \textit{image classification} approach requiring 8,193~KB of RAM (0.78\% of OPS-SAT capacity). More concerning is their deployment on CubeSat-class hardware, where the \textit{baseline} \textit{image classification} would consume over 50\% of available RAM, leaving insufficient memory for other critical spacecraft functions. Even the more modest \textit{direct classification} and \textit{forecasting \& threshold} approaches would claim approximately 10\% and 12\% of CubeSat RAM respectively---substantial allocations for a single subsystem.

Our optimization efforts dramatically transformed this landscape. The \textit{optimized} \textit{forecasting \& threshold} implementation now requires just 59 KB of RAM---a 34.6$\times$ reduction from its \textit{baseline}---representing a mere 0.36\% of CubeSat memory and an almost negligible 0.01\% of OPS-SAT resources. This exceptional efficiency preserves the highest detection performance (CEF$_{0.5}$ score of 88.8\%) among our \textit{optimized} approaches. Similarly, the \textit{optimized} \textit{direct classification} model achieved a 13.2$\times$ reduction in RAM usage to just 122~KB, requiring only 0.74\% of CubeSat memory. Both models demonstrate storage requirements well within spacecraft constraints, with Read-Only Memory (ROM) requirements of 166~KB and 149~KB respectively, each using less than 0.3\% of CubeSat storage capacity.

The \textit{image classification} approach, while significantly improved with a 8.0$\times$ reduction in RAM usage, still requires 1,024~KB (1 MB) of memory. While this represents a reasonable 6.25\% of CubeSat RAM, it remains substantially higher than our other approaches. This memory requirement, combined with its more intensive computational demands, makes \textit{image classification} less suitable for deployment on severely constrained platforms despite its impressive detection precision.

All three \textit{optimized} approaches easily meet the resource constraints of the OPS-SAT platform, with each requiring less than 0.1\% of available RAM and negligible portions of storage capacity. This indicates that more capable spacecraft computing platforms can readily support multiple concurrent anomaly detection systems, potentially enabling ensemble approaches that leverage the complementary strengths of different detection methods.

These results demonstrate that our neural architecture optimization approach successfully bridges the gap between sophisticated anomaly detection capabilities and the severe resource constraints of spacecraft computing environments. The \textit{optimized} \textit{forecasting \& threshold} and \textit{direct classification} models are particularly notable, as both satisfy common microcontroller deployment criteria ($<$250 KB RAM, $<$250 KB ROM) while requiring minimal computational resources. The \textit{forecasting \& threshold} approach represents the most balanced solution, with near-\textit{baseline} detection performance packed into an extraordinarily efficient implementation suitable for even the most constrained spacecraft computing platforms.

\begin{table*}[h!]
    \centering
    \caption{Resource Usage of Baseline and Optimized Anomaly Detection Systems Relative to On-Board Resources}
    \label{tab:resource_results}
    \begin{tabularx}{\textwidth}{l *{6}{>{\centering\arraybackslash}X}}
        \toprule
         & \multicolumn{3}{c}{Baseline} & \multicolumn{3}{c}{Optimized} \\
        \cmidrule(lr){2-4} \cmidrule(lr){5-7}
        Metric & \textit{Direct classification} & \textit{Image classification} & \textit{Forecasting \& threshold} & \textit{Direct classification} & \textit{Image classification} & \textit{Forecasting \& threshold} \\
        \midrule
        Event-wise F$_{0.5}$ (\%)     & 72.4 & 48.6 & \textbf{92.7} & 48.5 & 70.1 & 88.8 \\
        \midrule
        RAM (KB)      & 1,606   & 8,193    & 2,043   & 122    & 1,024   & \textbf{59}     \\
        ROM (KB)      & 1,268   & 4,902    & 1,294   & \textbf{149}    & 508     & 166    \\
        \midrule
        RAM (\%) CubeSat (16 MB)    & 9.80   & 50.01    & 12.47   & 0.74    & 6.25   & \textbf{0.36}     \\
        ROM (\%) CubeSat (64 MB)     & 1.93   & 7.48    & 1.97   & \textbf{0.23}    & 0.78     & 0.25    \\
        \midrule
        RAM (\%) OPS-SAT (1 GB)     & 0.15   & 0.78   & 0.19   & 0.01    & 0.10   & \textbf{0.01}     \\
        ROM (\%) OPS-SAT (8 GB)     & 0.02   & 0.06    & 0.02   & \textbf{$<$ 0.01}    & 0.01     & $<$ 0.01    \\
        \bottomrule
    \end{tabularx}
\end{table*}

Our optimization efforts yielded remarkable reductions in computational requirements across all three approaches, as shown in the ``\textit{optimized}'' columns of Table~\ref{tab:metric_results}. These \textit{optimized} models correspond to the solutions highlighted with green dots in Figures~\ref{fig:pareto_forecast}, \ref{fig:pareto_direct}, and \ref{fig:pareto_image}, representing our selected configurations from the respective \textit{Pareto fronts}.

The \textit{forecasting \& threshold} approach maintained the highest overall detection performance with a CEF$_{0.5}$ of 88.8\% while achieving dramatic resource reductions. Most notably, this \textit{optimized} implementation requires just 354k MACs---a 99.4\% reduction from its \textit{baseline} counterpart---while preserving near-perfect precision (99.9\%). Despite a moderate decrease in recall from 72.3\% to 61.6\%, the model maintained high operational utility while reducing its RAM footprint by 97.1\% to just 59 KB, making it deployable even on highly constrained spacecraft computing platforms.

The \textit{direct classification} approach experienced the most significant performance trade-off, with its CEF$_{0.5}$ decreasing from 72.4\% to 48.5\%. This substantial drop resulted primarily from reduced recall (from 36.9\% to 16.9\%), though precision remained relatively high at 91.1\%. This performance compromise may be attributed to two factors: first, our optimization targeted the CEF$_{0.5}$ metric which heavily weights precision over recall, potentially leading to threshold settings that favor precision; second, this approach underwent the most aggressive parameter reduction (16.4×), potentially limiting model capacity. However, this compromise enabled extraordinary computational efficiency gains, with parameters reduced to just 19K and MACs decreased by 98.4\% to 907k. The model's memory requirements shrank dramatically to just 122 KB of RAM and 149 KB of ROM---the smallest storage footprint among all \textit{optimized} implementations---making it potentially suitable for extremely resource-constrained deployment scenarios where detection of some anomalies is preferable to none.

Perhaps most surprising was the optimization of the \textit{image classification} approach, which not only reduced computational demands but actually improved detection performance. The CEF$_{0.5}$ increased from 48.6\% to 70.1\%, driven by a dramatic improvement in precision from 50.2\% to 99.1\%. While recall decreased moderately from 43.1\% to 32.3\%, the near-elimination of false positives represents a significant operational advantage. These improvements accompanied substantial resource reductions, with parameters decreased by 89.8\% to 128K and MACs reduced by 95.3\% to 45M. Despite these gains, the \textit{image classification} approach still requires significantly more computational resources than the other methods, although the near-elimination of false positives makes it worth considering for critical applications where false alarms are extremely costly.

These optimization results demonstrate that neural architecture optimization can dramatically reduce the computational requirements of spacecraft anomaly detection systems while maintaining operational utility. The \textit{optimized} \textit{forecasting \& threshold} implementation represents the most balanced solution, with high detection performance and minimal resource requirements. At just 59 KB of RAM and 354k MACs, this model is deployable on most spacecraft computing platforms, including severely constrained CubeSat systems. While the other approaches offer viable alternatives for specific operational contexts---\textit{direct classification} for absolute minimal resource usage and \textit{image classification} for extremely high precision---the \textit{optimized} \textit{forecasting \& threshold} approach provides the best overall combination of detection capability and computational efficiency for spacecraft telemetry anomaly detection.

\section{Conclusion and Future Work}
\label{Conclusion and Future Work}

\subsection{Conclusion}
This research demonstrates that sophisticated deep learning-based anomaly detection capabilities can be successfully deployed on space-grade edge devices through systematic neural architecture optimization. Our experiments with three distinct anomaly detection approaches---\textit{forecasting \& threshold}, \textit{direct classification}, and \textit{image classification}---reveal important trade-offs between detection performance and computational efficiency in the context of spacecraft telemetry analysis.

The \textit{forecasting \& threshold} approach consistently delivered the best balance of detection capability and resource efficiency, achieving a 92.7\% CEF$_{0.5}$ in its \textit{baseline} form and maintaining 88.8\% after optimization while reducing computational requirements by over 99\%. This approach represents a compelling solution for on-board anomaly detection, requiring just 59 KB of RAM (0.36\% of CubeSat memory) while maintaining near-perfect precision. However, its accuracy depends heavily on a sophisticated non-parametric dynamic thresholding mechanism that operates on batches of data, introducing computational overhead beyond the model itself.

The \textit{direct classification} approach offers unique advantages despite its lower CEF$_{0.5}$. As the most compact implementation at just 19K parameters and 149 KB ROM, it produces interpretable anomaly probabilities that can be converted to binary classifications with minimal post-processing. This characteristic, combined with its high precision (91.1\%) and ability to operate without batch-dependent thresholds, makes it particularly suitable for real-time detection in environments with limited computational capabilities.

While \textit{image classification} achieved competitive accuracy after optimization---increasing its CEF$_{0.5}$ from 48.6\% to 70.1\%---it comes with higher computational costs that limit its deployment options on the most constrained platforms. Nevertheless, its distinct approach offers complementary strengths that could be valuable in ensemble systems.

Our multi-objective optimization approach using the \textit{Pareto-optimal} methodology provides spacecraft engineers with a systematic framework for balancing detection performance against hardware constraints, enabling deployment of sophisticated anomaly detection even on platforms with severe computational limitations.

\subsection{Future Work}
Future research should explore several promising directions. First, investigating hybrid or ensemble methods that combine the complementary strengths of multiple detection approaches could potentially improve overall detection robustness without significantly increasing resource requirements. The high precision of \textit{forecasting \& threshold} coupled with the real-time processing advantages of \textit{direct classification} suggests particular promise in this area.

Second, addressing the remaining detection limitations in \textit{direct} and \textit{image classification} through more advanced thresholding mechanisms could further enhance performance without compromising computational efficiency. Additionally, extending our multi-objective optimization framework to include additional spacecraft-relevant objectives such as energy consumption and inference latency would provide a more comprehensive optimization landscape.

Third, comparing our neural architecture optimization approach with other model compression techniques like quantization, pruning, and knowledge distillation could reveal complementary pathways to efficiency. Scaling our solution to the full dataset of Mission 1 and Mission 2 would further generalize our findings across diverse spacecraft operational profiles.

Finally, integrating these anomaly detection systems with autonomous recovery and fault-management subsystems would enable comprehensive evaluation under realistic mission conditions, particularly through opportunities like ESA's OPS-SAT platform, providing invaluable validation of these approaches in authentic operational environments.

\FloatBarrier

\bibliographystyle{IEEEtran}
\bibliography{biblio}

\end{document}